\documentclass[lettersize,journal]{IEEEtran}
\usepackage{amsmath,amsfonts}
\usepackage{algorithm}
\usepackage{algorithmicx}
\usepackage{algpseudocode}
\usepackage{array}
\usepackage[caption=false,font=normalsize,labelfont=sf,textfont=sf]{subfig}
\usepackage{textcomp}
\usepackage{stfloats}
\usepackage{url}
\usepackage{verbatim}
\usepackage[dvipdfmx]{graphicx}
\usepackage{amssymb}
\usepackage[space, compress]{cite}
\usepackage{color}
\usepackage{enumitem}
\usepackage{amsmath,amssymb}
\usepackage{mathtools}
\usepackage{hyperref}

\begin{document}

\title{ViSA: Visited-State Augmentation \\for Generalized Goal-Space Contrastive Reinforcement Learning}

\author{
    Issa Nakamura$^{1*}$, 
    Tomoya Yamanokuchi$^{1*}$, 
    Yuki Kadokawa$^{1}$,\\ 
    Jia Qu$^{2}$, 
    Shun Otsubo$^{2}$, 
    Ken Miyamoto$^{2}$, 
    Shotaro Miwa$^{2}$ and Takamitsu Matsubara$^{1}$
    \thanks{
        *Equal contribution.
    }
    \thanks{
        $^{1}$I. Nakamura, T. Yamanokuchi, Y. Kadokawa, and T. Matsubara are with the Graduate School of Information Science, Nara Institute of Science and Technology (NAIST), Nara, Japan.{\tt\small}
    }%
    \thanks{
        $^{2}$J. Qu, S. Otsubo, Y. Susumu, S. Miwa are with the Advanced Technology R\&D Center, Mitsubishi Electric Corporation, Hyogo, Japan. {\tt\small}
    }%
}

\markboth{Preprint}
{Shell \MakeLowercase{\textit{et al.}}: A Sample Article Using IEEEtran.cls for IEEE Journals}

\maketitle

\begin{abstract}
    Goal-Conditioned Reinforcement Learning (GCRL) is a framework for learning a policy that can reach arbitrarily given goals. 
    In particular, Contrastive Reinforcement Learning (CRL) provides a framework for policy updates using an approximation of the value function estimated via contrastive learning, achieving higher sample efficiency compared to conventional methods.
    However, since CRL treats the visited state as a pseudo-goal during learning, it can accurately estimate the value function only for limited goals.
    To address this issue, we propose a novel data augmentation approach for CRL called ViSA (Visited-State Augmentation). ViSA consists of two components: 1) generating augmented state samples, with the aim of augmenting hard-to-visit state samples during on-policy exploration, and 2) learning consistent embedding space, which uses an augmented state as auxiliary information to regularize the embedding space by reformulating the objective function of the embedding space based on mutual information.
    We evaluate ViSA in simulation and real-world robotic tasks and show improved goal-space generalization, which permits accurate value estimation for hard-to-visit goals. 
    Further details can be found on the project page: \href{https://issa-n.github.io/projectPage_ViSA/}{\texttt{https://issa-n.github.io/projectPage\_ViSA/}}

\end{abstract}


\section{Introduction}
    \IEEEPARstart{G}{oal}-Conditioned Reinforcement Learning (GCRL)\cite{GCRL1,GCRL3,Robot_GCRL1,Robot_GCRL3} is a framework for learning a policy that can generate different actions depending on a given goal, and it has been widely used for multi-goal tasks. 
    Among GCRL methods, Contrastive Reinforcement Learning (CRL)\cite{Contrastive_RL} interprets GCRL as a goal-reaching problem and provides a framework for policy updates using an approximation of the value function estimated by Contrastive Learning (CL).
    It can achieve high sample efficiency even in robotic tasks with large goal spaces.
    Specifically, CRL encodes the goal reachability from a state-action pair (anchor) under the current policy as relative distances in the embedding space. In this space, states visited from the anchor during on-policy exploration (visited states) are pulled closer as pseudo-goals, while states sampled from different rollouts (random states) are pushed farther away.
    This is learned based on the on-policy rollouts and the empirical frequencies of reaching each state from the anchor.
    
    However, while on-policy exploration readily collects easy-to-visit states, states requiring specific conditions or multi-step procedures are rarely collected, leading to sample bias in the visited states used to learn the embedding space. As a result, the learned embedding space overfits to frequently visited states during on-policy exploration and cannot properly capture relative distances for hard-to-visit states. 
    Therefore, it may introduce estimation bias and fail to accurately estimate the value function for hard-to-visit goals.

    \begin{figure}[t]
      \centering
      \includegraphics[width=\linewidth]{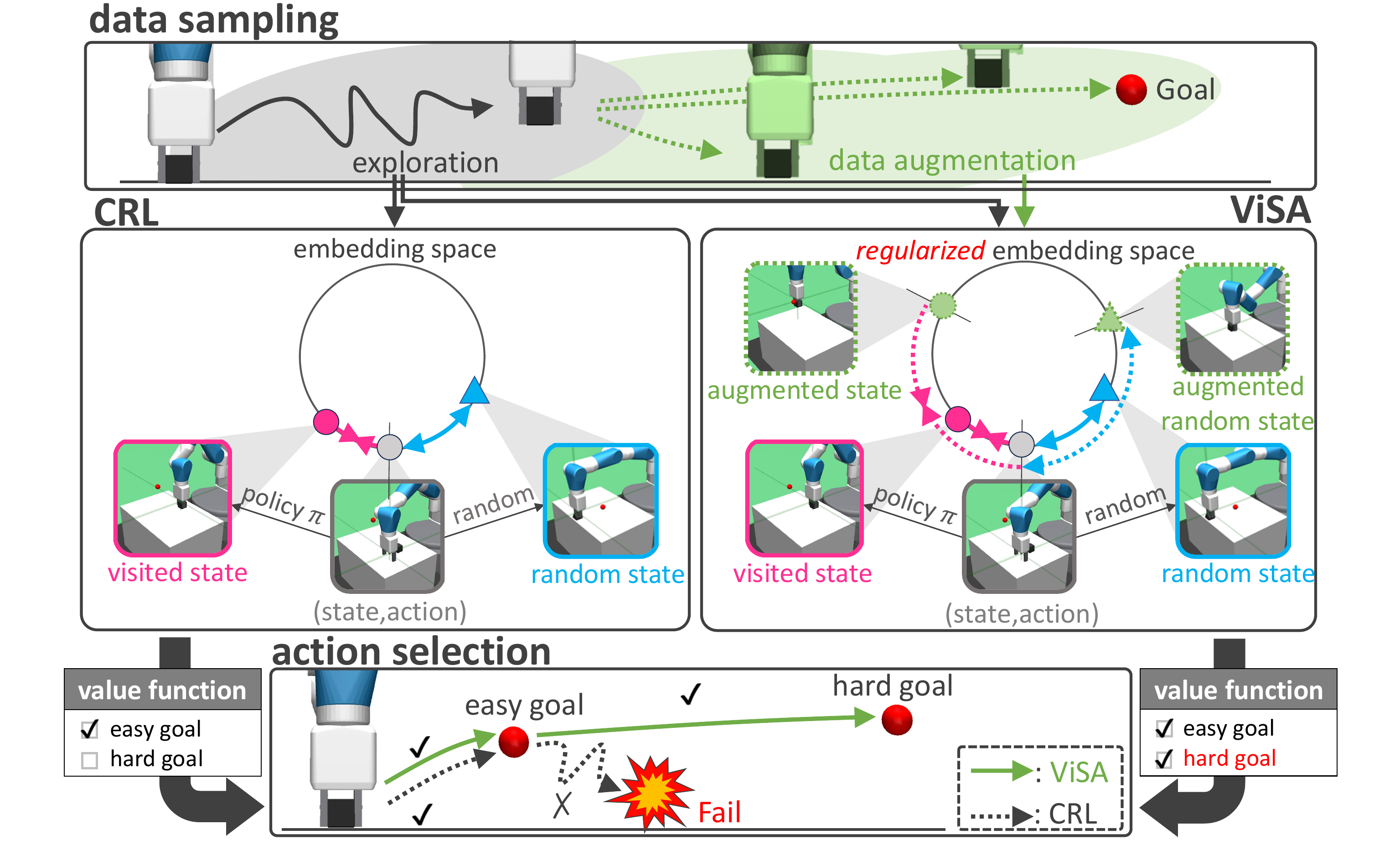}
      \vspace{-15pt}
      \caption{Overview of ViSA. This approach has two components: 1) generating augmented state samples to augment hard-to-visit state samples during on-policy exploration and 2) learning consistent embedding space, which uses an augmented state as auxiliary information to regularize the embedding space.
      The embedding space is trained to pull a visited state closer to a state-action pair (anchor) and push a random state farther away, encoding goal reachability from the anchor.
      ViSA suppresses estimation bias by considering relative distances of augmented states and augmented random states with respect to the anchor, enabling accurate value estimation and action selection for diverse goals in CRL.}
      \label{fig:Proposed_Method_Outline}
      \vspace{-15pt}
    \end{figure}
    
    To mitigate this bias, we artificially augment visited states (augmented states) that are \textit{inherently} reachable under current policy but insufficiently visited in limited on-policy rollouts.
    However, naive augmentation of visited state samples may lead to an inconsistent embedding space, where relative distances no longer correspond to goal reachability, due to introducing unreachable states or excessively oversampling rarely visited states.
    To address this issue, we draw inspiration from the embedding space regularization approaches developed in CL\cite{CL_PI1,CL_PI2}. 
    Instead of using augmented states merely as additional visited states, we leverage them as auxiliary information to regularize the embedding space, prompting the system to commonly calibrate relative distances to both rarely visited and originally visited states.
    As a result, this prevents overfitting to a limited set of visited states.
    
    In this paper, we propose Visited-State Augmentation (ViSA) as a novel data augmentation framework tailored to CRL  (Fig.~\ref{fig:Proposed_Method_Outline}).
    ViSA consists of two components: 1) generating augmented state samples and 2) learning consistent embedding space.
    In generating augmented state samples, we define an augmentation distribution conditioned on visited states and generate augmented state samples that include inherently reachable but rarely considered states in conventional on-policy rollouts.
    In learning consistent embedding space, we attempt generalization to large goal-spaces and more accurate value estimation by treating augmented states as auxiliary information for regularizing the embedding space.
    To achieve this, we first set as the objective function of the embedding space in CRL the application of mutual information between state-action pairs and visited states.
    Then, by applying mutual information factorization techniques\cite{Decomposed_MI_Estimation,Factorized_CL}, we reformulate the objective function using augmented states, while still preserving the definition of the original CRL objective function.
    In our study, the effectiveness of ViSA is evaluated through simulation and real-world experiments on multiple robotic tasks using a manipulator and a robotic hand, demonstrating that ViSA improves the goal-space generalization of CRL compared to the previous CRL method and GCRL baselines.
    
    The major contributions of this study can be summarized as follows:
    \begin{itemize}
        \item We propose a generalization approach to CRL that artificially augments visited state samples to address the sample bias issue inherent in the previous CRL method.
        \item To achieve consistent embedding space learning under the proposed data augmentation, we propose the ViSA framework to reformulate the objective function of the embedding space based on the mutual information factorization implemented to use augmented states as auxiliary information.
        \item We validate the effectiveness of ViSA in robotic task learning through simulation and real-world experiments.
    \end{itemize}

\section{Related Work}

    \subsection{Mitigating Sample Bias in GCRL}
        In GCRL for multi-goal tasks\cite{GCRL2,Robot_GCRL2}, sample bias induced by on-policy exploration is a common issue that degrades task performance. 
        To address this problem, several methods have been proposed to promote on-policy exploration of underexplored areas by modifying goal selection during exploration and training or by gradually adjusting goal difficulty based on obtained samples\cite{GDA_explore1,CHER,GDA_explore2,MHER}. 
        While these approaches mitigate sample bias by making effective use of limited samples, it remains challenging to collect samples from states that are rarely visited during on-policy exploration.
    
        CRL interprets GCRL as a goal-reaching problem and updates policy using an approximated value function via CL, permitting the reuse of rollouts across different goals and leveraging them to support value estimation. This design yields high sample efficiency, even in tasks with large goal spaces. 
        Several extensions have been developed to improve learning efficiency, including a TD-learning–based reformulation of CL-based value estimation for long-horizon multi-goal tasks \cite{TD_Contrasttive_RL}. 
        To enhance exploration capability, another method—although targeted at single-goal tasks—fixes the exploration goal to the final goal, thus reducing exploration variance and improving policy consistency and learning efficiency \cite{Expolation_Contrasttive_RL}.
        However, these approaches do not address the sample bias caused by reachability differences in the exploration of multi-goal tasks, leaving the problem of degraded accuracy in value estimation unresolved.

        In contrast to those prior studies, we propose a generalized CRL framework that directly mitigates the fundamental sample bias in on-policy CRL by augmenting hard-to-visit states, thus improving the accuracy of value estimation.  
    
    \subsection{Regularization of Embedded Spaces for CL}
        Contrastive learning is a representation learning framework that encodes dependencies among multiple random variables as relative distances in the embedding space. This approach has been applied widely in areas such as computer vision\cite{SimCRL,CL_image4,CL_image5} and time-series prediction\cite{TimeContrastive_related1,TimeContrastive_related2}. 
        However, CL is known to suffer from overfitting to observed samples. As a result, it often learns embedding spaces that capture task-irrelevant information or only partially task-relevant information, which can negatively affect downstream tasks\cite{CL_PI2,Factorized_CL}.
        To address this issue, several regularization approaches for embedding spaces\cite{CL_PI1, CL_PI2} have been proposed. These methods leverage auxiliary information sources to incorporate additional task-relevant information that is difficult to obtain from the original samples alone, and they use this information to adjust the structure of the embedding space.
       
        In contrast to previous CRL, we propose ViSA as a novel method that applies embedding space regularization techniques, well established in CL, to the value function estimation of CRL. Moreover, while conventional regularization methods in CL typically avoid augmented samples, since they can lead to inconsistent embedding spaces, ViSA leverages such samples by reformulating the embedding objective based on mutual information.

\section{Preliminaries}
    \subsection{Contrastive Learning}
    \label{PL: CL}
        As a representation learning framework, CL captures the conditional dependence between two random variables $x,y$ and encodes it as geometric distance in an embedding space.
        Formally, given their joint distribution $p(x,y)=p(y|x)p(x)$, we map observations of $x,y$ into embedding features $\psi(x),\phi(y)$ using encoders $\psi,\phi$, respectively. 
        The objective is to approximate the conditional distribution $p(y|x)$ through the contrastive critic $f(x,y)=\psi(x)^{T} \phi(y)$. 
        The learning data consist of anchor $x\sim p(x)$, positive example $y^+\sim p(y|x)$, and negative examples $y^-=\{y^{(1)},y^{(2)},...,y^{(K)}\}\sim p(y)$.
        CL is formulated as a conditional discrimination task in which the positive example $y^+$ should attain a higher similarity to anchor $x$ than to the negative examples, using the following InfoNCE loss:
        \begin{equation}
            I_{NCE}\left(x; y \mid \psi,\phi\right)
            = \mathbb{E}_{\substack{
            x\sim p(x)\\
            y^+\sim p(y|x)\\
            y^{(k)}\sim p(y)
            }}
            \left[
            \log \frac{e^{f(x,y^+)}}{\sum_{k=1}^K e^{f(x,y^{(k)})}}
            \right].
            \label{eqn: InfoNCE}
        \end{equation}
        The optimal contrastive critic $f^*$ obtained by maximizing Eq.~\eqref{eqn: InfoNCE} satisfies the following property:
        \begin{equation}
            \begin{split}
                f^*(x,y)=\log \frac{p(y|x)}{p(y)c(x)}.
                \label{eqn: Optimal Critic}
            \end{split}
        \end{equation}
        Here, $c(\cdot)$ is an arbitrary function. Since the InfoNCE loss is known to correspond to a lower bound on the mutual information $I(x;y)$ between $x$ and $y$ \cite{CPC}, the objective of CL can be interpreted as learning a contrastive critic $f$ that maximizes $I(x;y)$:
        \begin{equation}
            \begin{split}
                f^*(x,y)=\arg \max_{f} I(x;y)\approx \arg \max_{f} I_{NCE}(x;y).
                \label{eqn:MI_CL}
        \end{split}
        \end{equation}

    \subsection{Goal-Conditioned RL as a Goal-reaching Problem}
    \label{PL: GCRL}
        In a Goal-Conditioned Markov Decision Process (GCMDP), Goal-Conditioned RL (GCRL) is defined by six elements: $(\mathcal{S}, \mathcal{G}, \mathcal{A}, \mathcal{T}, r, and \gamma)$. Here, $\mathcal{S}$ is a set of the environment's observation space, $\mathcal{G}$ the set of goals, and $\mathcal{A}$ the action space. The state transition probability $\mathcal{T} = p(s' \mid s, a)$ is defined as the probability of transitioning to the next state $s' \in \mathcal{S}$ after taking action $a \in \mathcal{A}$ in state $s \in \mathcal{S}$. A goal $g \in \mathcal{G}$ corresponds to a subset of $\mathcal{S}$, $r(s, a, g)$ sets the reward for transitions toward the goal, and $\gamma \in [0,1)$ is the discount factor.

        In addressing goal-reaching problems, the objective of GCRL is to learn a policy $\pi(a | s, g)$ that can reach a given goal $g$. 
        Accordingly, the reward function $r$ is defined as the probability, and when it takes action $a$ in state $s$ following the current policy $\pi$, it reaches goal $g$ at the next state $s'$:
        \begin{eqnarray}
            \label{GCRL_reward}
            \begin{aligned}
                r(s, a, g) \triangleq (1-\gamma)p(s'=g|s,a).
            \end{aligned}
            \vspace{-10pt}
        \end{eqnarray}
        In this case, the goal-conditioned value function $Q^{\pi}(s, a, g)$ represents the cumulative reachability of the goal over future time steps under the current policy $\pi$. This can be defined as the discounted state-occupancy measure $p^{\pi}(s_{t+} \mid s, a)$, and it can be characterized via a Bellman formulation ~\cite{Contrastive_RL} as
        \begin{equation}
        \begin{split}
            Q^\pi(s, a, g)&\triangleq p^{\pi\left(\cdot \mid \cdot, g\right)}\left(s_{t+}=g \mid s, a\right)\\
            &=(1-\gamma) \sum_{t=0}^{\infty} \gamma^t p^{\pi\left(\cdot \mid \cdot, g\right)}\left(s_{t}=g \mid s, a\right).
        \end{split}
        \label{eqn: GCRL Q Function}
        \vspace{-10pt}
        \end{equation}
        Here, $s_{t+}$ denotes a state reached after the state-action pair $(s,a)$, and $s_t$ is the state reached exactly $t$ steps after $(s,a)$. The objective is to learn the optimal policy $\pi^*(a \mid s, g)$ that maximizes the optimal value function $Q^{*}(s,a,g)$ as follows:
        \begin{equation}
        \begin{split}
            \pi^*(a | s, g) = \arg \max_{\pi} Q^{*}(s, a, g).
        \end{split}
        \label{eqn: GCRL_policy}
        \vspace{-10pt}
        \end{equation}
        
    \subsection{Contrastive Reinforcement Learning}
    \label{subsec:preliminary:CRL}
        Contrastive reinforcement learning (CRL) estimates an approximation of the value function for goal-reaching problems using CL. Formally, the reachability of an arbitrary goal $g$ from a state-action pair $(s,a)$ under the current policy $\pi$ is encoded as relative distances in the embedding space.
        During learning, an arbitrary state-action pair $(s,a)$, which is collected from an on-policy rollout toward a given goal $g$, is used as an anchor. The state $s_v^{+}$ visited from the anchor is treated as a pseudo-goal and used as a positive example. Moreover, $K$ states $s_v^-=\{s_v^{(1)}, s_v^{(2)}, \ldots, s_v^{(K)}\}$ are randomly sampled from rollouts toward different goals from the anchor and used as negative examples.
        Based on the definition of the value function, the positive example is sampled as a future state visited $t$ steps after the anchor, where $t$ follows the geometric distribution \footnote{This geometric distribution is referred to as the visited state distribution.} $t\sim GEOM(1-\gamma)=(1-\gamma)\gamma^{t-1}$.

        Using these samples, the encoders $\psi,\phi$ extract embedding features $\psi(s,a),\phi(s_v)$ based on the InfoNCE loss:
        \begin{equation}
        \begin{split}
                &I_{NCE}((s,a);s_v)
                \\&= \mathbb{E}_{\substack{(s,a)\sim p(s,a)\\s_v^+\sim p(s_v|s,a)s_v^{(k)}\sim p(s_v)}}
                \left[\log \frac{e^{f(s, a, s_v^+)}}{\sum_{k=1}^K e^{f(s,a, s_v^{(k)})}}\right] 
                \\&\leq I((s,a);s_v).
                \label{eqn: CRL Loss is MI}
            \end{split}
            \vspace{-10pt}
            \end{equation}
        The optimal encoders $\psi^*,\phi^*$ maximize Eq.\eqref{eqn: CRL Loss is MI}, and the resulting optimal contrastive critic $f^*=\psi(\cdot)^{*T}\phi(\cdot)^*$ corresponds to the probability density ratio proportional to the reachability of the visited state $s_v$ from anchor $(s,a)$:
        \begin{equation}
            f^*(s,a,s_v)= \frac{p^{\pi\left(\cdot \mid \cdot, g\right)}(s_{t+}=s_v|s,a)}{p(s_v)c(s,a)}\propto Q^\pi(s, a,s_v).
            \label{eqn: CRL Real Q Function}
        \end{equation}
        As discussed in Section~\ref{PL: GCRL}, the goal-conditioned value function represents the discounted state-occupancy measure to the goal. Consequently, by replacing the goal $g$ with the visited state $s_v$, the optimal contrastive critic $f^*$ can be interpreted as an approximation of the value function $Q^{\pi}(s,a,g)$:
        \vspace{-5pt}
        \begin{equation}
            \begin{split}
                Q^\pi(s, a, g)&\triangleq f^*(s,a,s_v=g).
            \end{split}
            \label{eqn: CRL Q Function}
            \vspace{-5pt}
        \end{equation}
        As discussed in Section~\ref{PL: CL}, since InfoNCE loss can be formulated using mutual information, the objective of learning embedding space in CRL can be viewed as maximizing the mutual information $I((s,a);s_v)$ between anchor and visited state:
        \vspace{-5pt}
        \begin{equation}
            \begin{split}
                \psi^*,\phi^*=\arg \max_{\psi,\phi}
                I((s,a);s_v).
                \label{eqn:MI_objective_CRL}
        \end{split}
        \vspace{-5pt}
        \end{equation}
        The policy $\pi$ is updated based on Eq.~\eqref{eqn: GCRL_policy} to maximize the value function optimized through the above procedure.

    \begin{figure*}[t]
              \centering
              \includegraphics[width=\linewidth]{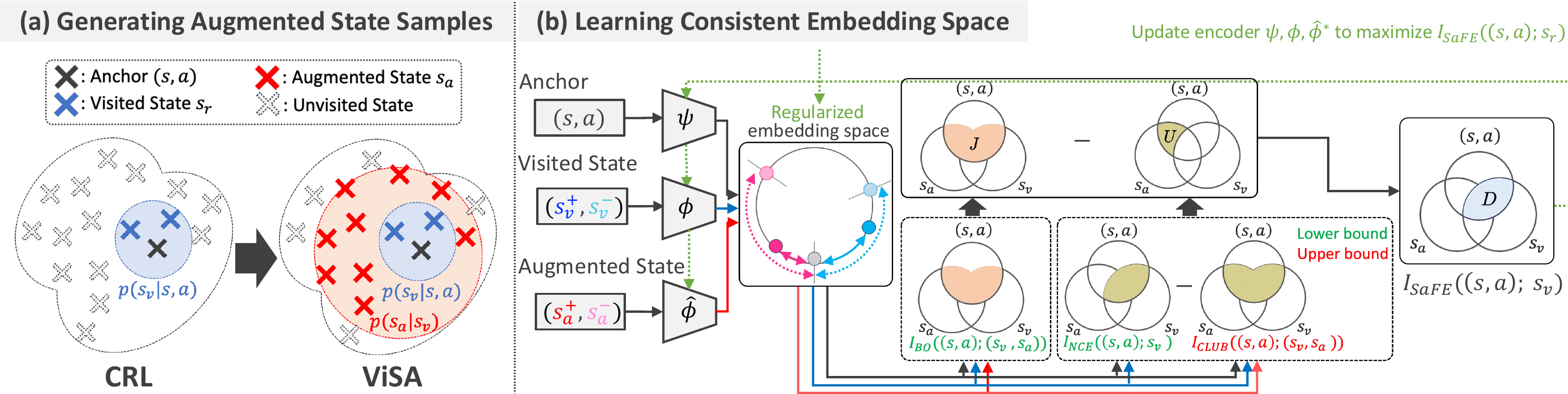}
              \caption{
              Framework of ViSA: (a) Generating augmented state samples. Using augmentation distribution $p(s_a \mid s_v)$, hard-to-visit states from on-policy rollouts are artificially augmented. Crosses indicate samples. Dotted region is inherently reachable state space, blue region contains samples from visited state distribution $p(s_v \mid s,a)$, and red region includes additional samples from $p(s_a \mid s_v)$.
              (b) Learning consistent embedding space. To regularize embedding space and prevent overfitting to visited states $s_v$, we use not only anchors $(s,a)$ and visited states $s_v^+, s_v^-$ from the previous CRL method but also augmented states $s_a^+, s_a^-$ as auxiliary information. Objective function of embedding space is reformulated based on  mutual information factorization, and encoders $\psi, \phi, \hat{\phi}$ are trained to maximize mutual information $I_{SaFE}((s,a); s_v)$ estimated using augmented states $s_a$. Mutual information terms inside dashed boxes are used to compute $I_{SaFE}((s,a); s_v)$, where red and green text indicate upper and lower bounds, respectively.}
              \label{fig:Critic Nerwork}
              \vspace{-10pt}
    \end{figure*}

\section{Proposed Method}
    ViSA is a novel data augmentation framework tailored to CRL, designed to achieve accurate value estimation for diverse goals.
    ViSA consists of two components: 1) generating augmented state samples and 2) learning consistent embedding space.
    In this section, we first explain the generation of augmented state samples.
    Next, we describe a reformulation of the objective function of the embedding space based on mutual information factorization, which allows the use of augmented states as auxiliary information.
    Finally, we explain the procedure used for learning consistent embedding space.
    
    \subsection{Generating Augmented State Samples}
        An overview of this component of the method is given in Fig.~\ref{fig:Critic Nerwork} (a).
        We define an augmentation distribution $p(s_a|s_v)$ conditioned on the visited states sampled during on-policy exploration and then generate augmented state $s_a$, which includes states that are inherently reachable but rarely visited in on-policy rollouts:
        \vspace{-10pt}
        \begin{equation}
            \begin{split}
                s_a \sim p(s_a|s_v).
                \label{eqn: Reachabel_State_Augmentation}
        \end{split}
        \vspace{-15pt}
        \end{equation}
        
        In principle, previous CRL  methods tend to select positive samples from visited state samples $s_v$ that are close to anchor $(s,a)$ when learning the embedding space, potentially leading to a shortage of samples distant from the anchor.
        To address this issue, we propose a simple method for defining the augmentation distribution $p(s_a|s_v)$, where augmented state $s_a$ is generated from the same rollout as the visited state $s_v$.
        Specifically, we define the augmentation distribution $p(s_a|s_v)$ as in Eq.~\eqref{eqn: GEOM_Reverse}, such that states visited near the end of an episode are selected with higher probability.
        Here, $t$ is time step sampled from the geometric distribution, and augmented states $s_a^+,s_a^-$ are generated as the states visited $t$ steps after the visited states $s_v^+,s_v^-$, respectively.
        \begin{equation}
            \begin{split}
                s_a^+ &\sim p(s_a=s_t|s_v=s_v^+), s_a^- \sim p(s_a=s_t|s_v=s_v^-)\\
                t&\sim 1-GEOM(1-\gamma)=1-(1-\gamma)\gamma^{t-1}
                \label{eqn: GEOM_Reverse}
        \end{split}
        \vspace{-10pt}
        \end{equation}
        
    \subsection{Reformulation of Objective Function}
        \label{Sec:PM2}
        To use augmented states as auxiliary information for learning the embedding space, we reformulate the objective function based on mutual information factorization \cite{Factorized_CL,Decomposed_MI_Estimation}. 
        In the field of CL, it is known that leveraging auxiliary information can regularize the embedding space \cite{CL_PI1,CL_PI2}, thus preventing overfitting to partially learned samples.
        Drawing inspiration from those works, we treat augmented state $s_a$ as auxiliary information and handle it as a distinct variable separate from visited state $s_v$ sampled during on-policy exploration.
        Then, by applying the factorized rule of mutual information to Eq.~\eqref{eqn: CRL Loss is MI}, it becomes possible to reformulate the objective function via the augmented state $s_a$:
        \begin{equation}
            \begin{split}
                \underbrace{I((s,a);s_v)}_{D}&=\underbrace{I((s,a);(s_v,s_a))}_{J} - \underbrace{I((s,a);s_a|s_v)}_{U}.
                \label{eqn: FCRL MI Decomposition1}
        \end{split}
        \vspace{-10pt}
        \end{equation}
        Here, $D$ denotes the mutual information representing reachability from the anchor to visited state $s_v$, corresponding to the objective function of the previous CRL method. $J$ is the joint mutual information that encompasses reachability to both $s_v$ and $s_a$, and $U$ is the unique mutual information that overestimates reachability to $s_a$.
        Accordingly, our method uses the mutual information $I_{SaFE}$ (Source-augmented Factorized Estimation) as the objective function of the embedding space:
        \vspace{-5pt}
        \begin{equation}
            \begin{split}
                I_{SaFE}&\triangleq J-U.
                \label{eqn:MI_SaFE}
        \end{split}
        \vspace{-5pt}
        \end{equation}

        This allows us to use augmented state $s_a$ as auxiliary information while preserving the original objective definition of the embedding space in CRL.
        
    \subsection{Learning Consistent Embedding Space}
        Our method regularizes the embedding space to prevent overfitting to visited state samples $s_v$ during on-policy exploration.
        Here, using the encoders $\psi,\phi,\hat{\phi}$, we extract
        the embedding features $\psi(s,a), \phi(s_v), \hat{\phi}(s_v,s_a)$, respectively.
        Then, the objective of our method is to train optimal encoders $\psi^*,\phi^*,\hat{\phi}^*$ that maximize the objective function of embedding space $I_{SaFE}$, described in section \ref{Sec:PM2}.
        \begin{equation}
            \begin{split}
                \psi^*,\phi^*,\hat{\phi}^*=\arg \max_{\psi,\phi,\hat{\phi}} I_{SaFE}
                \label{eqn:Optimized_MI_SaFE}
        \end{split}
        \vspace{-5pt}
        \end{equation}
        In the following, we describe the procedure for estimating joint mutual information $J$ in Section \ref{method:Joint_MI_J} and unique mutual information $U$ in section \ref{method:Unnecessary_MI_U}. 
        An overview of how we learn consistent embedding space is given in Fig.~\ref{fig:Critic Nerwork} (b).
        
        \subsubsection{\textbf{Estimation of Joint Mutual Information $J$}}
        \label{method:Joint_MI_J}
        Joint mutual information $J$ is estimated by computing a lower bound of the mutual information between the anchor $(s,a)$ and the pair of visited and augmented states $(s_v, s_a)$ by referring to Boosted Critic Estimation $I_{BO}$: \cite{Decomposed_MI_Estimation}:
        \begin{equation}
            \begin{split}
                &J \leq I_{BO}((s,a); (s_v,s_a) | f_{v,a}). 
            \end{split}
            \label{eqn: FCRL Joint Value}
            \vspace{-10pt}
        \end{equation}
        Here, we define the contrastive critic $h$ that approximates the reachability of visited and augmented states from the anchor as follows:
        \vspace{-5pt}
        \begin{equation}
            f_{v,a}(s,a,s_v,s_a) =\psi(s,a)^T \phi(s_v)+\psi(s,a)^T \hat{\phi}(s_v,s_a).
        \label{eqn: feature2val}
        \end{equation}

        In CL, the amount of information that can be compared at once is limited \cite{InfoNCE_is_bound_by_batch}, making the estimation of joint mutual information, which is generally richer information, prone to instability. 
        However, by using $I_{BO}$, the information can be factorized and compared separately, which stabilizes the learning process \cite{Decomposed_MI_Estimation}.

    \subsubsection{\textbf{Estimation of Unique Mutual Information $U$}}
        \label{method:Unnecessary_MI_U}
        By definition, in CRL it is impossible to collect anchor samples $(s,a)$ conditioned on visited state $s_v$, and thus the conditional mutual information conditioned to $s_v$ cannot be estimated directly.
        To address this, we estimate unique mutual information $U$ by reapplying the factorization rule of mutual information:
        \begin{equation}
            U \approx I_{NCE}
            ((s,a);s_v|f_v) - I_{CLUB}((s,a);(s_v,s_a)|f_{v,a}). 
            \label{eqn: FCRL Conditional MI}
        \end{equation}
        Here, we define the contrastive critic $l$ that approximates the reachability of a visited state from the anchor, as follows:
        \begin{equation}
            f_v(s,a,s_v)=\psi(s,a)^T \phi(s_v).
            \label{eqn:feature_1val}
        \end{equation}
        Moreover, to efficiently retrieve reachability information under the current policy while discarding unnecessary information, we simultaneously maximize the lower bound and minimize the upper bound of the mutual information, following Factorized Contrastive Learning~\cite{Factorized_CL}. 
        Specifically, we estimate the lower bound of mutual information $I_{NCE}((s,a);s_v|f_v)$ using InfoNCE loss~\cite{CPC} in Eq.~\eqref{eqn: CRL Loss is MI} and the upper bound $I_{CLUB}((s,a);(s_v,s_a)|f_{v,a})$ using CLUB loss~\cite{Factorized_CL}. 
        
        Based on the discussions above, we summarize the overall learning procedure of ViSA in Alg.~\ref{alg1}.

    \floatname{algorithm}{Alg.}
    \algtext*{EndWhile}
    \algtext*{EndFor}
    \begin{algorithm}[t]
    \caption{Learning Procedure of ViSA}
    \label{alg1}
    \begin{algorithmic}[0]
    \setlength{\baselineskip}{0.9\baselineskip}
    \Require Dataset {$(s,a)$, $s_v^+$, $s_v^-$} 
    \State Initialize encoders $\psi$, $\phi$, $\hat{\phi}$
    \While{}
      \State Sampling dataset $\{(s,a), s_v^+, s_v^-\}$ from replay buffer
      \For{sample batch \{$(s,a)$, $s_v^+, s_v^-$\}}
        \State Generate $s_a^+$ from Eq.~\eqref{eqn: GEOM_Reverse} based on $s_v^+$
        \State Generate $s_a^-$ from Eq.~\eqref{eqn: GEOM_Reverse} based on $s_v^-$
        \State Estimate $J:I_{BO}((s,a);(s_v,s_a))$ from Eq.~\eqref{eqn: FCRL Joint Value}
        \State Estimate $U:I_{NCE}((s,a);s_v)$ and 
        \State \hspace{1.9cm}$I_{CLUB}((s,a);(s_v,s_a))$ from Eq.~\eqref{eqn: FCRL Conditional MI}
        \State Calculate $I_{SaFE}=D = J-U$ from Eq.~\eqref{eqn:MI_SaFE}
        \State Update $\psi$, $\phi$, $\hat{\phi}$ to maximize $I_{SaFE}$
        \State Update $\pi$ to maximize $Q^{\pi}(s,a,g)$ from Eq.~\eqref{eqn: GCRL_policy}
      \EndFor
    \EndWhile
    \end{algorithmic}
    \end{algorithm}
        
\section{Experiments}
    \subsection{Experimental Purpose}
        To evaluate the effectiveness of our method, we conducted comparative experiments against baselines as well as analytical experiments. Through these experiments, we aim to answer the following four questions \footnote{The common experimental setting and an additional analysis of the effect of the relevance between visited and augmented states on mutual information estimation accuracy are shown in the project page: \href{https://issa-n.github.io/projectPage_ViSA/}{\texttt{https://issa-n.github.io/projectPage\_ViSA/}}}.
        \renewcommand{\labelenumi}{RQ\theenumi)}
        \begin{enumerate}[leftmargin=4em]
            \item Does our method improve the learning performance of CRL across a variety of robotic tasks? (\ref{subsec:ex4})
            \item Does our method outperform other sample diversification methods? (\ref{subsec:ex2})
            \item Does the effectiveness of our method depend on the choice of the augmentation distribution? (\ref{subsec:ex3})
            \item Is our method effective for real-world application? (\ref{subsec:ex5})
        \end{enumerate}
        \vspace{-9pt}

    \subsection{Simulation Learning of Robot Tasks}
    \label{subsec:ex4} 
    \begin{figure}[t]
              \centering
              \captionsetup{skip=-2pt}
              \includegraphics[width=\linewidth]{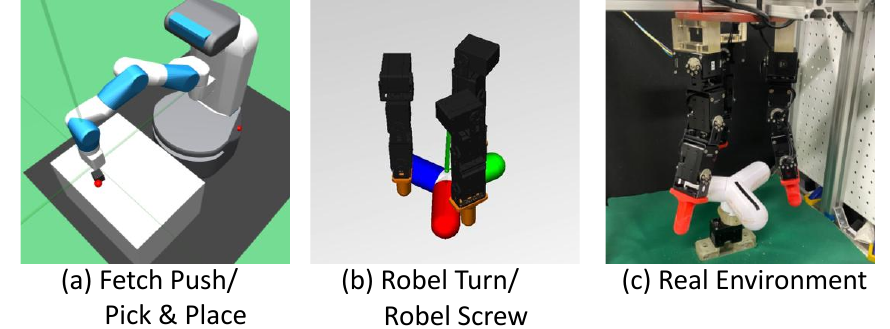}
              \caption{Experimental environments}
              \label{fig:Simulator}
              \vspace{-10pt}
        \end{figure}
    
    \subsubsection{\textbf{Setup}}
        To evaluate the proposed method's effectiveness for learning robotic tasks (\textbf{RQ1}), we conducted simulation learning with six robotic tasks and compared their task performance. The simulators used are shown in Fig.~\ref{fig:Simulator} (a) and (b), and an overview of the tasks is provided below.
        \begin{itemize}[leftmargin=2em]
            \item \textbf{Fetch Push\cite{Contrastive_RL}}: A task in which a manipulator pushes a box to the target position on a table.
            \item \textbf{Fetch Push Hard}: A variant of \textbf{Fetch Push} with an additional constraint requiring the box to be moved along the shortest path.
            \item \textbf{Pick \& Place\cite{HER}}: A task in which a manipulator grasps a box and transports it to the target position on a table or in the air.
            \item \textbf{Pick \& Place Hard}: A variant of \textbf{Pick \& Place} with an additional constraint on maintaining the box height.
            \item \textbf{Robel Turn\cite{Robel}}: A task in which a three-fingered robotic hand rotates a valve to an arbitrary target angle.
            \item \textbf{Robel Screw\cite{Robel}}: A task in which the robot rotates a valve by half a turn at a constant speed.
        \end{itemize}
        
        In addition, for Fetch Push and Pick \& Place tasks, the initial positions of the end-effector and the box, as well as the goal positions, were uniformly randomized. For Robel Turn, the initial valve positions, robot fingertips, and goals were uniformly randomized, whereas for Robel Screw, the robot fingertips were uniformly randomized but the valve was initialized to one of four discrete positions.

        As methods for comparison, we used two CRL variants, one employing BinaryNCE loss (CRL (NCE)) and the other InfoNCE loss (CRL (CPC)) \cite{Contrastive_RL}, C-Learning \cite{C_Learning}, which bootstraps value estimates from reward signals as does the GCRL baseline, and Goal-Conditioned Behavioral Cloning (GCBC) \cite{GCBC}, which performs imitation learning from randomly collected exploratory trajectories.

        \begin{figure}[t]
            \centering
            \captionsetup{skip=-2pt}
            \includegraphics[width=\linewidth]{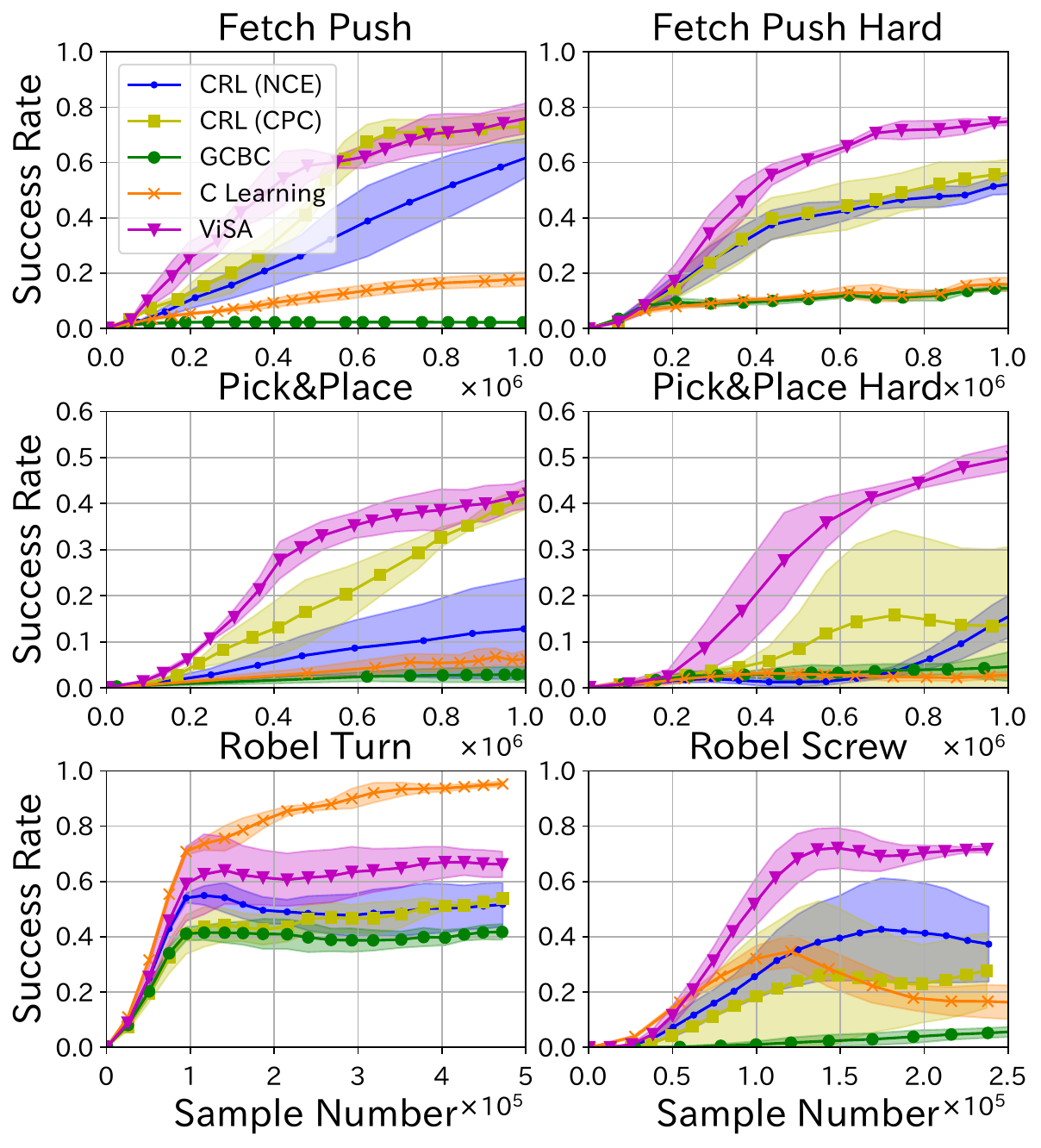}
            \caption{Learning results from robot tasks. Plots show success rate during learning for each task. Solid lines indicate the mean over three trials, and shaded areas represent variance.}
            \label{fig:Result Robot Task} 
            \vspace{-10pt}
        \end{figure}
        
    \subsubsection{\textbf{Results}}
        The learning curves of success rates are shown in Fig.~\ref{fig:Result Robot Task}.
        Overall, our method (solid purple lines) generally achieves task performance comparable to or better than the previous CRL methods.
        Moreover, in difficult task variants, where consideration of hard-to-sample visited state samples during on-policy rollouts is critical for task achievement, our method learned more efficiently and achieved higher success rates, as shown by the comparisons between \textbf{Fetch Push} and \textbf{Fetch Push Hard} and between \textbf{Pick \& Place} and \textbf{Pick \& Place Hard}.

        Therefore, these results suggest that our method is effective for learning robotic tasks, particularly under conditions where on-policy exploration tends to induce significant sample bias.
    
    \subsection{Comparison with Naive Sample Diversification Methods}
    \label{subsec:ex2}
        
    \subsubsection{\textbf{Setup}}
        If our purpose were merely to mitigate sample bias by diversifying visited state samples, we could adopt a naive diversification strategy within a previous CRL method without using our method.
        For example, this can be achieved by adjusting the discount factor $\gamma$, which controls the spread of the visited state distribution, and by more broadly collecting positive examples from the visited state samples.

        However, in this experiment, to answer \textbf{RQ2}, we compared our method and a naive CRL sample diversification method that varies discount factor $\gamma$ of the visited state distribution.
        Specifically, we conducted learning with three different discount factors $\gamma \in \{0.99, 0.999, 0.9999\}$.
        The visited state distribution for each compared method is shown in Fig.~\ref{fig:result_Ex2} (a).
        As the learning task, we used the Robel Screw task.
        
    \subsubsection{\textbf{Results}}
\begin{figure}[t]
            \centering
            \captionsetup{skip=-2pt}
            \includegraphics[width=\linewidth]{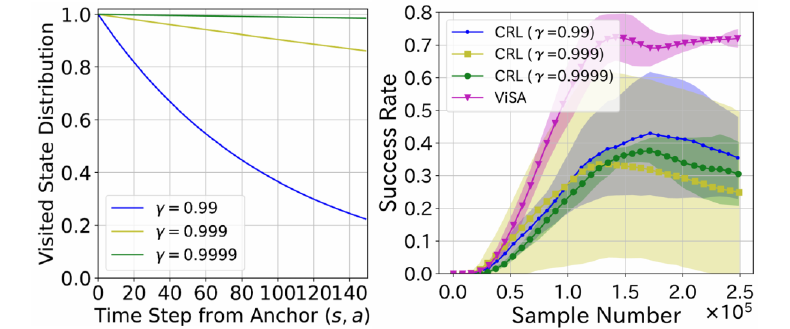}
            \caption{Visited state distributions and learning results for a naive sample diversification CRL method.
            (a) Visualization of visited state distributions. CRL is modified by adjusting discount factor $\gamma$ to collect visited state samples $s_v$ more broadly from on-policy rollouts. For visualization, sampling probabilities are normalized so that the maximum value is 1.
            (b) Learning curves of success rates. Solid lines show mean success rates over three trials for proposed method and baselines, and shaded regions designate variance.}
            \label{fig:result_Ex2} 
            \vspace{-10pt}
        \end{figure}

        The learning curves of success rates for our method and a naive CRL sample diversification method are shown in Fig.~\ref{fig:result_Ex2}~(b).
        CRL methods that adjust discount factors, i.e., $\gamma=0.999$ (solid yellow line) and $\gamma=0.9999$ (solid green line), exhibit large variance across trials and slightly worse performance than the original setting before changing $\gamma$ (solid blue line).
        This is likely because, although adjusting $\gamma$ increases the diversity of visited state samples, it reduces the difference in sample frequencies across time steps from the anchor, thus blurring the relative value ordering determined by the number of steps to reach the goal.
        In contrast, our method of using augmented states $s_a$ as auxiliary information (solid purple line) preserves the relative ordering of the value function according to the time steps required to reach the visited state
        in the embedding space, leading to faster convergence and lower variance across trials and thus more stable learning.

        Therefore, a naive CRL sample diversification method is inadequate for improving task performance, whereas using augmented states $s_a$ as auxiliary information to learn a consistent embedding space leads to improved performance.

    \subsection{Ablation Experiments with Augmentation Distributions}
    \label{subsec:ex3}
    \subsubsection{\textbf{Setup}}
\begin{figure}[t]
            \centering
            \captionsetup{skip=-2pt}
            \includegraphics[width=\linewidth]{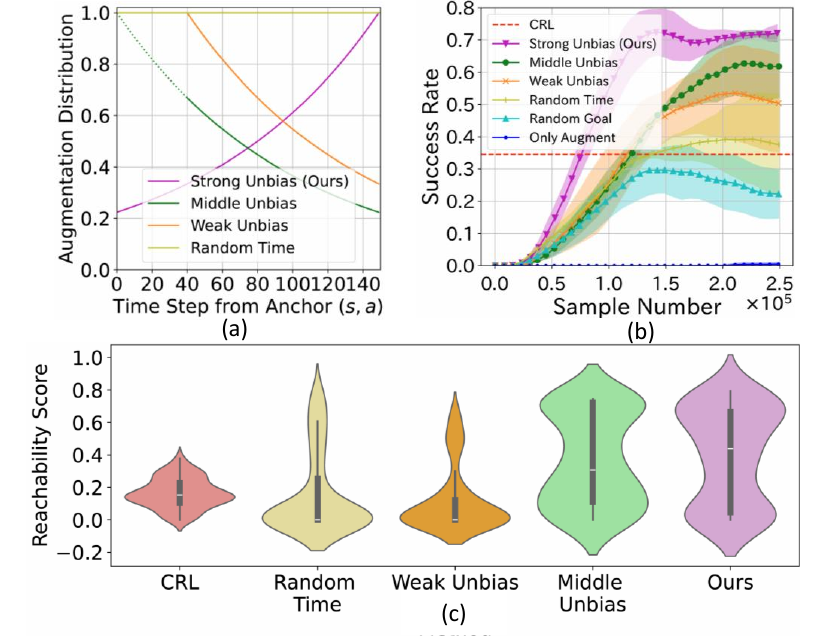}
            \caption{Augmentation distributions and learning results for ablations with different generation ranges of augmented states $s_a$:
            (a) Visualization of augmentation distributions. Each distribution is heuristically defined by adjusting time steps from the anchor that are likely to be selected as augmented states. For visualization, sampling probabilities are normalized so that the maximum value is 1.
            (b) Learning curves of success rates. Dashed red line denotes maximum success rate achieved by previous CRL method, while solid lines show mean success rates over three runs for each method. Shaded regions indicate variance.
            (c) Visualization of sample coverage with respect to reachability for visited and augmented states. Black lines show range from minimum to maximum values, and shaded regions represent sample distribution. Values closer to 1 correspond to hard-to-visit states requiring more steps, whereas values closer to 0 indicate easy-to-visit states.}
            \label{fig:result_Ex3} 
        \end{figure}

        In our method, augmentation distribution must be specified to generate augmented states $s_a$.
        However, the amount of information on hard-to-visit states provided by $s_a$ to complement the visited state $s_v$ is changed, and its effect on on-policy learning has not been sufficiently clarified, depending on how this distribution is defined.
        To answer \textbf{RQ3}, we conducted an experiment using seven ablation methods with different augmentation distributions.
        We evaluated these ablations by learning the Robel Screw task.
        The augmentation distribution used in each ablation is shown in Fig.~\ref{fig:result_Ex3} (a), and the details are described as follows.
        \begin{itemize}
            \item 
                \textbf{Strong Unbias (proposed)}: 
                States visited near the end of an episode are augmented more frequently to prioritize hard-to-\textit{sample} states in previous CRL methods.

            \item 
                \textbf{Weak Unbias}: 
                Future states relative to visited state $s_v$ are augmented by masking the visited state distribution in the previous CRL method, with states closer to $s_v$ sampled more frequently.
            \item 
                \textbf{Middle Unbias}: Future states after visited state sample $s_v$ are augmented. While states closer to $s_v$ are generated with higher probability, a wider range of timesteps is covered compared to \textbf{Weak Unbias}.
            \item
                \textbf{Random Time}: Augmented states $s_a$ are sampled uniformly from the same rollout as visited state $s_v$, and states visited before $s_v$ may be included.
            \item 
                \textbf{Random Goal}: 
                Augmented states $s_a$ are sampled randomly from the replay buffer and may have weak relevance to the visited state $s_v$, since they are collected from rollouts toward different goals.
            \item 
                \textbf{Only Augment}: Augmented states $s_a$ are generated as \textbf{Strong Unbias} but are not used as auxiliary information. Joint mutual information $J$ is used as the objective function of embedding space.
        \end{itemize}

        In addition, to quantitatively evaluate the effect of different augmentation distributions on sample diversity, we visualized sample coverage corresponding to reachability.
        Specifically, letting $T$ denote the episode length and $N$ the number of steps by which a visited state is sampled in approaching the goal from the anchor, we define $N/T$ as the reachability score used to measure sample coverage.
        For a fair comparison, the number of visited state samples in the previous CRL method is set equal to the total number of visited state and augmented state samples used in our method.

    \subsubsection{\textbf{Results}}
        The results of the ablation experiment with different augmentation distributions $p(s_a \mid s_v)$ are shown in Fig.~\ref{fig:result_Ex3} (b).
        Among the ablations, our method (\textbf{Strong Unbias}) achieved the highest success rate. While the final performance gradually decreased in the order of \textbf{Middle Unbias}, \textbf{Weak Unbias}, and \textbf{Random Time}, all of these ablations outperformed the previous CRL method.
        We attribute this improvement to augmenting visited states near the end of episodes, which permits learning a regularized embedding space over more diverse samples.
        In contrast, \textbf{Random Goal}, which generates augmented state samples $s_a$ weakly related to the visited state $s_v$, and \textbf{Only Augment}, which does not reformulate the objective function, performed worse than the previous CRL.
        These results confirm that learning performance does not improve when augmented states are generated randomly or when they are not used as auxiliary information.
        
        The sample coverage of visited and augmented states is shown in Fig.~\ref{fig:result_Ex3} (c). This shows that augmenting visited states that are hard to sample with the previous CRL method effectively expands sample coverage. In particular, \textbf{Strong Unbias (proposed)} and \textbf{Middle Unbias}, which yielded substantial performance improvement, exhibited higher reachability scores. This indicates that many hard-to-visit states are augmented and that sample coverage is significantly expanded.

        Therefore, designing the augmentation distribution $p(s_a \mid s_v)$ to preserve relevance to the visited state while intensively augmenting states that are hard to visit in on-policy exploration is crucial for improving both task performance and sample coverage.
        
    \subsection{Real-world Learning}
    \label{subsec:ex5}
        \begin{figure}[t]
            \centering
            \includegraphics[width=\linewidth]{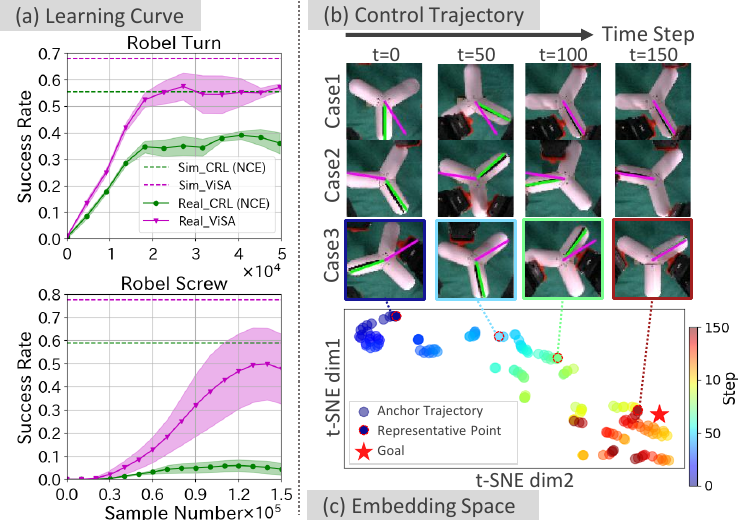}
            \caption{Real-world learning results:
            (a) Learning curves of success rates. Dashed lines denote maximum success rates achieved in simulation for each method, and solid lines show success rates during real-robot learning.
            (b) Control trajectories for Robel Turn using the policy trained by proposed method. Cases 1–3 correspond to trajectories toward different goals. Black, green, and purple lines indicate true valve angle, measured angle, and goal angle, respectively.
            (c) t-SNE visualization of embedding space. Each circle represents a state-action pair, dotted circles denote representative points corresponding to the snapshot, stars denote goals, and colors indicate time steps. Distances between circles and stars reflect value, with shorter distances indicating easy-to-reach states (higher value).}   
            \label{fig:result_Ex5} 
        \end{figure}
    \subsubsection{\textbf{Setup}}
       To evaluate the effectiveness of our method for real-world robot learning (\textbf{RQ4}), we conducted comparative experiments against the previous CRL method by using the Robel robot (Fig.~\ref{fig:Simulator} (c)) to evaluate the Robel Turn and Robel Screw tasks. 
       Each task was randomized under the same settings as in the simulation described in Section~\ref{subsec:ex4}.
       In addition, for Robel Turn, which has a large exploration-space, we initialized the policy with a network trained in simulation (Section~\ref{subsec:ex4}) and fine-tuned it on the real robot.
    \subsubsection{\textbf{Results}}        
        The learning curves of success rates on the real Robel robot are shown in Fig.~\ref{fig:result_Ex5} (a), and the valve control trajectories in Robel Turn are given in Fig.~\ref{fig:result_Ex5} (b).
        The results indicate that for Robel Turn, both our method and the previous CRL method made progress in learning, although performance slightly decreased compared to simulation. This is likely due to the small discrepancies between the true valve angle (black line) and the measured values (green line) in Fig.~\ref{fig:result_Ex5} (b), which arise from increased state diversity.
        Nevertheless, the final success rate of our method converged around 60\%, which is approximately 1.5 times higher than that of the previous CRL method. A similar trend was observed for the Robel Screw, with even more pronounced performance differences when learning completely from scratch.

        For Robel Turn, we visualized the embedding space using the trained encoder from our method. For each rollout, embedding features were extracted using the goal targeted during on-policy exploration and all state-action pairs as anchors, and then they were compressed into two dimensions using t-SNE. The resulting visualization is shown in Fig.~\ref{fig:result_Ex5} (c).
        In our method, embedding features of temporally adjacent state-action pairs are positioned close to each other and thus form a smooth trajectory toward the goal. This confirms that our method can estimate values according to both the reachability between states and that to the goal.

        These results indicate that ViSA is effective in real-robot learning and enhances the goal-space generalization of CRL.
\section{Discussion}
    In this section, we discuss the limitations of our method.
    The first limitation is that designing the augmentation distribution requires heuristic adjustments. In this work, augmented states were generated using manually tuned distributions, which is labor-intensive and task-dependent. A future direction is to automate this process, for example by adding a decoding function to the embedding space to generate augmented states that maintain their relevance to on-policy exploration data while introducing randomized variations \cite{FCL_D1,FCL_D2}.
    
    The second limitation is that our experiments used only a single augmentation distribution. However, the current framework could support multiple distributions by adding new encoders for each, at the cost of increasing network size. Future work includes efficiently generating augmented states using multiple distributions and sharing encoders across all distributions \cite{Decomposed_MI_Estimation} to keep the network simple.

\section{Conclusion}
    In this study, we proposed Visited-State Augmentation (ViSA), a novel data augmentation framework tailored to CRL, to achieve accurate value estimation over diverse goals. ViSA consists of two components: 1) generating augmented state samples and 2) learning consistent embedding space. Through both simulated and real-world robot task learning, we demonstrated that ViSA improves goal-space generalization in CRL and that embedding space regularization is crucial for enhancing task performance.

\bibliographystyle{ieeetr}
\bibliography{reference}

@inproceedings{Robel, 
  title        = {{ROBEL:} Robotics Benchmarks for Learning with Low-Cost Robots},
  author       = {Michael Ahn and
                  Henry Zhu and
                  Kristian Hartikainen and
                  Hugo Ponte and
                  Abhishek Gupta and
                  Sergey Levine and
                  Vikash Kumar},
  booktitle    = {Conf. Robot Learn.},
  pages        = {1300--1313},
  year={2019}
}

@inproceedings{Contrastive_RL, 
  title={Contrastive learning as goal-conditioned reinforcement learning},
  author={Eysenbach, Benjamin and Zhang, Tianjun and Levine, Sergey and Salakhutdinov, Russ R},
  booktitle={Adv. Neural Inf. Process. Syst.},
  volume={35},
  pages={35603--35620},
  year={2022}
}

@inproceedings{TD_Contrasttive_RL,
  author       = {Chongyi Zheng and
                  Ruslan Salakhutdinov and
                  Benjamin Eysenbach},
  title        = {Contrastive Difference Predictive Coding},
  booktitle    = {Int. Conf. Learn. Represent.},
  pages = {47577--47601},
  year         = {2024},
}

@inproceedings{Expolation_Contrasttive_RL,
  author       = {Grace Liu and
                  Michael Tang and
                  Benjamin Eysenbach},
  title        = {A Single Goal is All You Need: Skills and Exploration                         Emerge from Contrastive {RL} without Rewards,                               Demonstrations, or Subgoals},
  booktitle    = {Int. Conf. Learn. Represent.},
  year         = {2025},
}

@inproceedings{CL_PI1,
  author       = {Xin Yuan and 
                  Zhe L. Lin and 
                  Jason Kuen and 
                  Jianming Zhang and 
                  others},
  title        = {Multimodal Contrastive Training for Visual Representation Learning},
  booktitle    = {Comput. Vis. Pattern Recognit.},
  pages        = {6995--7004},
  year         = {2021},
}

@article{CL_PI2,
title={Unsupervised Domain Adaptation by Learning Using Privileged Information},
author={Adam Breitholtz and Anton Matsson and Fredrik D. Johansson},
journal={Trans. Mach. Learn. Res.},
issn={2835-8856},
note={ISSN: 2835-8856},
volume       = {2024},
year={2024},
}

@inproceedings{Robot_GCRL1,
  title={Sim-to-real transfer of robotic control with dynamics randomization},
  author={Peng, Xue Bin and Andrychowicz, Marcin and Zaremba, Wojciech and Abbeel, Pieter},
  booktitle={IEEE Int. Conf. Robot. Autom.},
  pages={3803--3810},
  year={2018},
}

@inproceedings{Robot_GCRL2,
  title={Nonprehensile planar manipulation through reinforcement learning with multimodal categorical exploration},
  author={Ferrandis, Juan Del Aguila and Moura, Jo{\~a}o and Vijayakumar, Sethu},
  booktitle={Int. Conf. Intell. Robots Syst.},
  pages={5606--5613},
  year={2023},
}

@article{Robot_GCRL3,
  title={Goal-conditioned reinforcement learning with disentanglement-based reachability planning},
  author={Qian, Zhifeng and You, Mingyu and Zhou, Hongjun and Xu, Xuanhui and He, Bin},
  journal={Robot. Autom. Lett.},
  volume={8},
  number={8},
  pages={4721--4728},
  year={2023},
}

@inproceedings{GCRL1,
  title={Generalizing goal-conditioned reinforcement learning with variational causal reasoning},
  author={Ding, Wenhao and Lin, Haohong and Li, Bo and Zhao, Ding},
  booktitle={Adv. Neural Inf. Process. Syst.},
  volume={35},
  pages={26532--26548},
  year={2022}
}

@inproceedings{GCRL2,
  title={Goal-conditioned reinforcement learning with imagined subgoals},
  author={Chane-Sane, Elliot and Schmid, Cordelia and Laptev, Ivan},
  booktitle={Int. Conf. Mach. Learn.},
  pages={1430--1440},
  year={2021},
}

@inproceedings{GCRL3,
  author       = {Soroush Nasiriany and
                  Vitchyr Pong and
                  Steven Lin and
                  Sergey Levine},
  title        = {Planning with Goal-Conditioned Policies},
  booktitle    = {Adv. Neural Inf. Process. Syst.},
  volume={32},
  pages        = {14814--14825},
  year         = {2019},
}

@article{GDA_explore1,
  title={Goal exploration augmentation via pre-trained skills for sparse-reward long-horizon goal-conditioned reinforcement learning},
  author={Wu, Lisheng and Chen, Ke},
  journal={Mach. Learn.},
  volume={113},
  number={5},
  pages={2527--2557},
  year={2024},
}

@article{GDA_explore2,
  title={Goal-conditioned on-policy reinforcement learning},
  author={Gong, Xudong and 
          Dawei, Feng and 
          Xu, Kele and 
          Ding, Bo and 
          others},
  journal={Adv. Neural Inf. Process. Syst.},
  volume={37},
  pages={45975--46001},
  year={2024}
}

@inproceedings{Decomposed_MI_Estimation,
    author = {Sordoni, Alessandro and
              Dziri, Nouha and
              Schulz, Hannes and
              Gordon, Geoff and
              others},
    title = {Decomposed Mutual Information Estimation for Contrastive Representation Learning},
    booktitle = {Int. Conf. Mach. Learn.},
    pages = {9859--9869},
    year = {2021},
}

@inproceedings{Factorized_CL,
    title={Factorized Contrastive Learning: Going Beyond Multi-view Redundancy},
    author={Liang, Paul Pu and
            Deng, Zihao and
            Ma, Martin and
            Zou, James and
            others},
    booktitle={Adv. Neural Inf. Process. Syst.},
    volume = {36},
    pages = {32971--32998},
    year={2023}
}

@inproceedings{HER,
  author       = {Marcin Andrychowicz and
                  Dwight Crow and
                  Alex Ray and
                  Jonas Schneider and
                  others},
  title        = {Hindsight Experience Replay},
  booktitle    = {Neural Inf. Process. Syst.},
  pages        = {5048--5058},
  year         = {2017},
}

@inproceedings{C_Learning,
  author       = {Benjamin Eysenbach and
                  Ruslan Salakhutdinov and
                  Sergey Levine},
  title        = {C-Learning: Learning to Achieve Goals via Recursive Classification},
  booktitle    = {Int. Conf. Learn. Represent.},
  year         = {2021},
}

@inproceedings{MHER,
  author       = {Rui Yang and
                  Meng Fang and
                  Lei Han and
                  Yali Du and
                  others},
  title        = {{MHER:} Model-based Hindsight Experience Replay},
  booktitle    = {Deep RL Workshop at Neural Inf. Process. Syst.},
  year         = {2021},
}

@inproceedings{CHER,
  author       = {Meng Fang and
                  Tianyi Zhou and
                  Yali Du and
                  Lei Han and
                  others},
  title        = {Curriculum-guided Hindsight Experience Replay},
  booktitle    = {Adv. Neural Inf. Process. Syst.},
  vol          = {32},
  pages        = {12602--12613},
  year         = {2019},
}

@article{CL_image4,
  title={Debiased contrastive learning},
  author={Chuang, Ching-Yao and
          Robinson, Joshua and
          Lin, Yen-Chen and
          Torralba, Antonio and
          others},
  journal={Adv. Neural Inf. Process. Syst.},
  pages={8765--8775},
  year={2020}
}

@article{CL_image5,
  title={Supervised contrastive learning},
  author={Khosla, Prannay and 
            Teterwak, Piotr and 
            Wang, Chen and 
            Sarna, Aaron and 
            others},
  journal={Adv. Neural Inf. Process. Syst.},
  pages={18661--18673},
  year={2020}
}

@inproceedings{TimeContrastive_related1, 
  author       = {Jean{-}Yves Franceschi and
                  Aymeric Dieuleveut and
                  Martin Jaggi},
  title        = {Unsupervised Scalable Representation Learning for Multivariate Time
                  Series},
  booktitle    = {Adv. Neural Inf. Process. Syst.},
  pages        = {4652--4663},
  year         = {2019},
}

@inproceedings{TimeContrastive_related2, 
  author       = {Seunghan Lee and
                  Taeyoung Park and
                  Kibok Lee},
  title        = {Soft Contrastive Learning for Time Series},
  booktitle    = {Int. Conf. Learn. Represent.},
  year         = {2024},
}

@inproceedings{SimCRL,
  author       = {Ting Chen and
                  Simon Kornblith and
                  Mohammad Norouzi and
                  Geoffrey E. Hinton},
  title        = {A Simple Framework for Contrastive Learning of Visual Representations},
  booktitle    = {Int. Conf. Mach. Learn.},
  pages        = {1597--1607},
  year         = {2020},
}

@article{CPC,
  author       = {A{\"{a}}ron van den Oord and
                  Yazhe Li and
                  Oriol Vinyals},
  title        = {Representation Learning with Contrastive Predictive Coding},
  journal      = {Comput. Res. Repos.},
  eprinttype   = {arXiv},
  eprint       = {1807.03748},
  year         = {2018},
}

@inproceedings{InfoNCE_is_bound_by_batch,
    author       = {Ben Poole and
                  Sherjil Ozair and
                  A{\"{a}}ron van den Oord and
                  Alexander A. Alemi and
                  others},
    title        = {On Variational Bounds of Mutual        
                    Information},
    booktitle      = {Int. Conf. Mach. Learn.},
    pages        = {5171--5180},
    year         = {2019}, 
}

@inproceedings{FCL_D1,
  title={What to align in multimodal contrastive learning?},
  author={Benoit Dufumier and
                  Javiera Castillo Navarro and
                  Devis Tuia and
                  Jean{-}Philippe Thiran},
booktitle={Int. Conf. Learn. Represent.},
  volume={},
  pages={},
  year={2025},
}

@inproceedings{FCL_D2,
  title={A Contrastive Learning Approach for Training Variational Autoencoder Priors},
  author={Jyoti Aneja and
                  Alexander G. Schwing and
                  Jan Kautz and
                  Arash Vahdat},
  booktitle={Adv. Neural Inf. Process. Syst.},
  pages={480--493},
  year={2021},
}

@inproceedings{GCBC,
  title={Decision Transformer: Reinforcement Learning via Sequence Modeling},
  author={Lili Chen and
                  Kevin Lu and
                  Aravind Rajeswaran and
                  Kimin Lee and
                  Aditya Grover and
                  others},
  booktitle={Adv. Neural Inf. Process. Syst.},
  pages={15084--15097},
  year={2021},
}

\end{document}